\documentclass[runningheads]{llncs}

\usepackage[T1]{fontenc}
\usepackage{graphicx}
\usepackage{multirow}
\usepackage{url}
\usepackage{bbding}
\usepackage{orcidlink}

\hypersetup{hidelinks}

\begin{document}

\title{CMNIE: An Information Extraction Benchmark for Chinese Military News}
\titlerunning{CMNIE}

\author{Yan Yu$^{\orcidlink{0009-0005-7409-8383}}$ \and Mengna Zhu \and Zhenyu Song \and Hao Yang \and  \\
Haiwen Chen \and Mao Wang\textsuperscript{(\Envelope)}}
\authorrunning{Y. Yu et al.}
\institute{Laboratory for Big Data and Decision, National University of Defense Technology, Changsha, China\\
\email{\{yanyu, zhumengna16, songzhenyu\}@nudt.edu.cn\\
\{yanghao21, chenhaiwen13, wangmao\}@nudt.edu.cn
}}

\maketitle

\begin{abstract}
Structured extraction from Chinese military news supports intelligence analysis, decision-making, and knowledge base construction.
However, existing resources provide limited support for joint information extraction in this domain, especially when events, event arguments, entities, and relations must be modeled together.
We present CMNIE, an information extraction benchmark for Chinese military news.
Extending military-domain resources beyond document-level event annotations, CMNIE jointly annotates event triggers, event arguments, named entities, and entity relations under a unified domain schema.
The dataset contains 13,000 instances collected from public Chinese military news, with manual annotations for 7 event types, 10 argument roles, 7 entity types, and 8 relation types.
We evaluate supervised IE models, zero-shot large language models, and fine-tuned LLM-based extraction methods on a shared test set.
Experimental results show that CMNIE remains challenging, especially for relation extraction and exact matching of event-argument spans; zero-shot LLMs often identify relevant semantic units but fail to match gold span boundaries exactly.
CMNIE provides a standardized benchmark for studying schema adherence, exact span matching, and joint structured extraction in specialized Chinese news.

\end{abstract}

\section{Introduction}

Information extraction (IE) from military news supports intelligence analysis, decision support, and structured knowledge base construction by converting unstructured reports into structured records of entities, relations, events, and event arguments.
More broadly, structured event information can facilitate downstream event-centric multi-document summarization \cite{zhu-etal-2025-eventsum} and decision-oriented reasoning over complex event contexts \cite{zhu-etal-2026-large}.
Progress in IE has been closely tied to benchmark datasets, including ACE 2005 \cite{ace2005}, DuIE \cite{liDuIELargescaleChinese2019}, DuEE \cite{li-etal-2020-duee}, and the MAVEN series \cite{wang-etal-2022-maven,wang-etal-2024-maven,wang-etal-2020-maven}.
These datasets support reliable comparison and provide partial coverage of Chinese military news, where event descriptions, equipment names, organizations, locations, and entity relations occur in domain-specific, information-dense contexts.

\begin{figure}[tbp]
\centering
\includegraphics[width=0.85\textwidth]{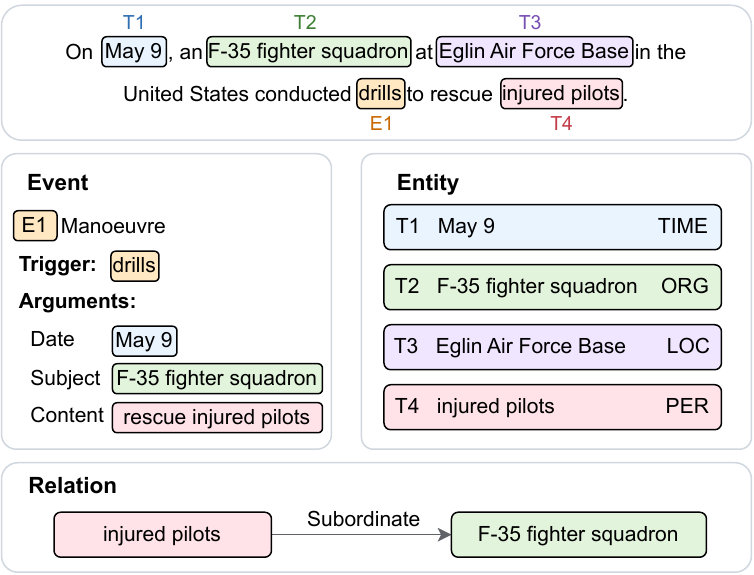}
\caption{Example of the CMNIE information extraction task and annotation layers.}
\label{fig:annotation-example}
\end{figure}

Military news poses several challenges for IE.
It contains long or abbreviated mentions of weapons, military units, facilities, geopolitical entities, and organizations.
Event arguments may be distant from the trigger or embedded in descriptive clauses, and relations such as ownership, subordination, cooperation, and hostility often appear without explicit verbal predicates.
A single instance may also contain entities, relations, and event arguments that depend on one another.
A benchmark that aligns all four layers enables direct measurement of whether a system can recover the full information structure of an instance.
Figure \ref{fig:annotation-example} illustrates a translated CMNIE instance and the four annotation layers used by the benchmark.

The closest military-domain resource is CMNEE, a document-level Chinese military event extraction dataset \cite{zhu-etal-2024-cmnee}.
Its annotations center on event triggers and arguments, leaving joint evaluation of events, entities, and entity relations in Chinese military news underexplored.

This paper presents CMNIE as a resource and benchmark for Chinese military news IE, with a unified annotation schema covering events, arguments, entities, and relations.
The benchmark is designed for both conventional supervised IE systems and recent extraction methods based on large language models (LLMs).
In addition to exact-match F1, we report relaxed event scores for zero-shot LLMs because generative models often locate the correct semantic unit but return a boundary that differs from the reference span.

The contributions are as follows.
First, we introduce CMNIE, a Chinese military news IE benchmark with joint annotations for events, event arguments, named entities, and entity relations.
Second, we describe the construction of CMNIE, including source collection, instance construction, schema design, human annotation, adjudication, and quality assessment.
Third, we provide benchmark results for supervised IE models, zero-shot LLMs, and fine-tuned LLM-based extraction methods, together with analyses of span matching and major error types.

\section{Related Work}

IE benchmarks differ in both task scope and annotation unit.
ACE 2005 defines multilingual entity, relation, and event annotations \cite{ace2005}; DuIE provides large-scale Chinese relation triples \cite{liDuIELargescaleChinese2019}; and DuEE covers 65 event types and 121 argument roles in real-world Chinese news \cite{li-etal-2020-duee}.
WikiEvents supports document-level event argument extraction, while the MAVEN series expands from event detection to event relations and exhaustive argument annotation \cite{li2021document,wang-etal-2022-maven,wang-etal-2024-maven,wang-etal-2020-maven}.
Together, these resources show the value of connected annotation layers, while their general-domain schemas and annotation units leave specialized Chinese military information structures only partially represented.
Domain-specific IE resources align their schemas with local terminology and information needs.
CMNEE is the closest resource to CMNIE and provides document-level triggers and arguments for Chinese military events \cite{zhu-etal-2024-cmnee}.
CMNIE uses an independently collected corpus and aligns all four layers through a schema developed from its own corpus analysis and annotation discussions.
Table \ref{tab:comparison} compares CMNIE with representative IE resources in terms of instance count, event mentions, event labels, argument roles, entity types, and relation types \cite{li2021document,ace2005,wang-etal-2024-maven,zhu-etal-2024-cmnee}.

\begin{table}[h]
\caption{Comparison with representative information extraction datasets.}
\label{tab:comparison}
\centering
\small
\setlength{\tabcolsep}{2pt}
\begin{tabular}{l r r r r r r}
\hline
Dataset & Instances & \shortstack{Event\\Mentions} & \shortstack{Event\\Labels} & \shortstack{Argument\\Roles} & \shortstack{Entity\\Types} & \shortstack{Relation\\Types} \\
\hline
ACE 2005 & 599 & 4,090 & 33 & 36 & 45 & 18 \\
WikiEvents & 246 & 3,951 & 50 & 59 & -- & -- \\
MAVEN-ARG & 4,480 & 98,912 & 162 & 612 & -- & -- \\
CMNEE & 17,000 & 29,223 & 8 & 11 & -- & -- \\
CMNIE & 13,000 & 6,997 & 7 & 10 & 7 & 8 \\
\hline
\end{tabular}
\end{table}

IE methods likewise vary in how they represent spans and schemas.
DyGIE++ uses contextualized span representations, OneIE performs globally informed joint extraction, and UIE casts multiple IE tasks as unified structure generation \cite{lin-2020-joint,lu-2022-unified,wadden-2019-entity}.
Recent LLM-based methods encode task specifications more explicitly: GoLLIE conditions generation on annotation guidelines, ADELIE aligns LLMs with IE supervision, and KnowCoder represents schemas as code \cite{li-etal-2024-knowcoder,qi-2024-adelie,sainz2024gollie}.
Zero-shot LLMs further test whether general instruction following transfers to specification-heavy extraction with strict label and boundary conventions \cite{peng2023doesincontextlearningfall}.

\section{CMNIE Dataset}

\subsection{Source Collection and Instance Construction}

CMNIE was constructed from public Chinese military news sources with high domain relevance.
We collected raw news pages from Sina Military\footnote{\url{https://mil.news.sina.com.cn/}}, Phoenix Military\footnote{\url{https://mil.ifeng.com/}}, and NetEase Military\footnote{\url{https://war.163.com/}}.
Using Python Requests and BeautifulSoup, we crawled more than 40,000 military news articles published from August 2010 to February 2022 in chronological order.
After crawling, duplicate pages, malformed entries, navigation text, HTML fragments, garbled text, and other boilerplate were removed; using three sources reduces dependence on a single editorial source.

To obtain domain-relevant candidate instances, we first built a candidate trigger dictionary.
Starting from the predefined event inventory, an LLM was used only to suggest synonyms and related expressions; all suggested terms were manually screened before dictionary matching.
Dictionary matching retained about 20,000 candidate texts; manual filtering removed about 7,000 and retained 13,000 for annotation.
Manual screening applied three criteria: topical relevance required the instance to fall within the CMNIE schema; textual quality required complete, readable, nonduplicate contexts free of webpage noise; and annotation tractability favored instances with a manageable number of independent events so that all four layers could be annotated consistently.
The dictionary was used only for candidate retrieval; all final labels and spans were manually annotated from the complete retained context.
The retained set includes instances with and without event annotations; after topical and quality screening, length sorting served only to organize the manual review sequence.

Chinese word segmentation was used to support annotation and checking.
The texts were segmented with the Harbin Institute of Technology LTP toolkit \cite{che-etal-2010-ltp}.
To reduce segmentation errors for military terminology, we collected domain terms before segmentation and added them to the segmentation dictionary.
The final annotations are span based; segmentation serves as a workflow aid, with every boundary anchored directly to the source text.

\subsection{Annotation Schema}

The schema was constructed through corpus analysis, pilot annotation, and expert discussion.
The initial event inventory was derived from the target military news domain and revised by checking whether triggers and arguments could be consistently identified; this process led to the addition of Exhibit, which frequently appeared in equipment display and public demonstration news.

CMNIE defines four annotation layers: event labels, event argument roles, entity types, and relation types.
Each event-bearing instance contains one trigger span, one Event Type, and zero or more event arguments with Argument Roles; other instances have empty event annotations.
The initial inventory contained six event types: Experiment, Manoeuvre, Deploy, Accident, Indemnity, and Support.
During annotation, many instances described a subject displaying equipment at a time or place, so Exhibit was added as a seventh event type.
Indemnity is retained as the dataset event type for logistics and supply support events.

Table \ref{tab:schema} summarizes the event label definitions and valid argument roles.
All type, role, and relation names are kept consistent with the dataset annotation.
The argument roles are Subject, Equipment, Date, Location, Content, Area, Militaryforce, Result, Object, and Materials.
The entity types are PER, TIME, GPE, EQU, ORG, LOC, and FAC.
The relation types are Belong, Carry, Subordinate, Equal, Hostility, Develop, Cooperation, and Fellow.

\begin{table}[t]
\caption{CMNIE event labels and argument role constraints.}
\label{tab:schema}
\centering
\small
\setlength{\tabcolsep}{2pt}
\begin{tabular}{p{1.7cm} p{4.9cm} p{5.1cm}}
\hline
Event Label & Definition & Argument Roles \\
\hline
Experiment & Tests or verifies equipment. & Subject, Equipment, Date, Location \\
Manoeuvre & Describes an exercise, drill, or training activity. & Subject, Content, Date, Area \\
Deploy & Moves or stations forces or equipment. & Subject, Militaryforce, Date, Location \\
Accident & Reports an unexpected harmful incident. & Subject, Result, Date, Location \\
Indemnity & Provides logistics, supplies, or service support. & Subject, Object, Materials, Date \\
Support & Provides assistance or relief. & Subject, Object, Date \\
Exhibit & Displays or publicizes equipment. & Subject, Equipment, Date, Location \\
\hline
\end{tabular}
\end{table}

Entity annotation captures the main mention types that fill event roles and relation arguments.
PER covers individuals and groups of people; TIME includes dates, times, and temporal spans; and GPE covers countries, cities, provinces, states, and other political entities.
EQU encompasses weapons, platforms, systems, and other military equipment, while ORG includes military units, institutions, companies, and other organizations.
LOC is used for non-political locations, bases, regions, and other place expressions; FAC is reserved for functional structures and infrastructure, including airports, buildings, railways, and bridges.

Relation annotation captures stable or contextual links between entity mentions.
Belong records ownership or operational association between equipment and an organization or geopolitical entity.
Carry represents a carrier--carried configuration involving a person or another piece of equipment, while Subordinate encodes hierarchical affiliation to a higher organization or geopolitical entity.
Equal connects mentions of the same real-world entity; Hostility records adversarial links; Develop captures development links; Cooperation represents collaborative links; and Fellow denotes peer relations.
The Equal relation is important because military news frequently alternates between full equipment names and shortened mentions.
Event arguments and named entities are separate annotation layers, so a longer argument span may intentionally contain a more specific entity span, as illustrated by the Content argument and the nested entity in Figure \ref{fig:annotation-example}.

\subsection{Annotation Workflow and Quality Assessment}

\begin{figure}[!htbp]
\centering
\includegraphics[width=0.95\textwidth]{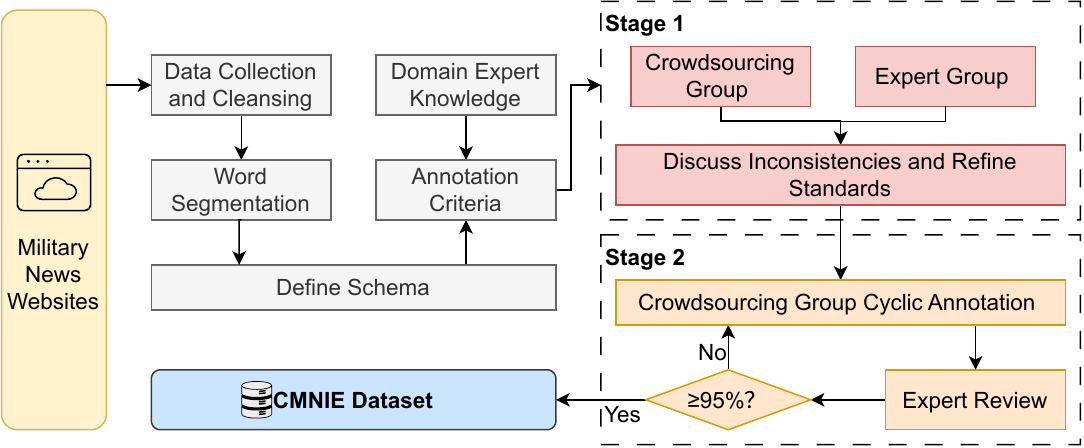}
\caption{Construction and annotation workflow of CMNIE.}
\label{fig:construction-workflow}
\end{figure}

The annotation workflow followed a two-stage, multi-round design adapted to the four CMNIE annotation layers.
Automatic event type pre-labels were used only as references; annotators manually judged event types, trigger spans, argument spans and roles, entity spans and types, and relation triples.
The annotation team included 20 experienced general annotators, all of whom received CMNIE-specific training, and five domain experts responsible for guideline development, difficult cases, and disagreement review.
In the first stage, experienced annotators and domain experts independently annotated shuffled batches, and disagreements were reviewed after each batch to refine boundary rules, role constraints, and relation examples.
In the second stage, annotator teams corrected all batches according to the finalized standard, while domain experts audited sampled instances and resolved difficult cases until the batch-level quality indicators reached the required thresholds.
Figure \ref{fig:construction-workflow} summarizes the data construction and annotation workflow.

For quality assessment, we used four audit indicators: event type accuracy, event argument recall, named entity recall, and relation recall.
A batch was accepted only when all four indicators reached at least 95\%; otherwise, it was returned for discussion and correction.
The final audit scores were 95.3\%, 95.1\%, 99.4\%, and 97.8\%, respectively. These values measure final audit quality after adjudication.

To quantify annotation reliability before adjudication, ten general annotators independently annotated the same 500 instances.
Table \ref{tab:agreement} reports the resulting pre-adjudication inter-annotator agreement (IAA): Fleiss' kappa for event annotation status and type (empty or one of the seven event types), and mean pairwise exact-match F1 for the structured layers, requiring all relevant spans and labels to match.

\begin{table}[!htbp]
\caption{Pre-adjudication IAA on 500 CMNIE instances independently annotated by ten annotators.}
\label{tab:agreement}
\centering
\small
\setlength{\tabcolsep}{4pt}
\begin{tabular}{l l r}
\hline
Annotation Layer & Agreement Measure & Result \\
\hline
Event annotation: empty or one of seven types & Fleiss' kappa & 0.9075 \\
Entity span and entity type & Pairwise exact-match F1 & 0.8688 \\
Trigger span and event type & Pairwise exact-match F1 & 0.8741 \\
Relation endpoints and relation type & Pairwise exact-match F1 & 0.7266 \\
Argument span, event type, and role & Pairwise exact-match F1 & 0.7463 \\
\hline
\end{tabular}
\end{table}

Event-type agreement is highest; relations and arguments are lower because they combine span and label decisions and inherit component disagreements. This pattern supports cross-checking and expert adjudication.

\subsection{Dataset Statistics}

\begin{figure}[t]
\centering
\includegraphics[width=1\textwidth]{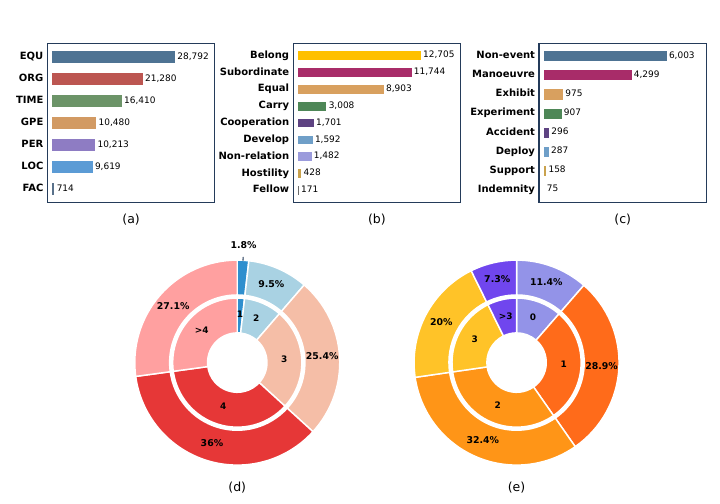}
\caption{Dataset statistics of CMNIE. (a) Entity types. (b) Relation annotations, including empty annotations. (c) Event annotations, including empty annotations. (d) Entities per instance. (e) Relations per instance.}
\label{fig:dataset-distribution}
\end{figure}

CMNIE contains 13,000 news instances.
Among them, 6,997 contain event annotations and 6,003 have empty event annotations.
Because every event instance contains one trigger span, the dataset contains 6,997 event triggers.
The dataset contains 23,087 event arguments, 97,508 named entity mentions, and 40,252 relation mentions.
The final training, development, and test splits were produced at the instance level.
A greedy split strategy was used to keep event-label distributions as similar as possible across the three subsets.
Table \ref{tab:stats} reports split statistics, and Figure \ref{fig:dataset-distribution} shows the main type distributions and instance-level density patterns.

\begin{table}[!htbp]
\caption{CMNIE split statistics.}
\label{tab:stats}
\centering
\small
\setlength{\tabcolsep}{2.6pt}
\begin{tabular}{l r r r r r r}
\hline
Split & Instances & Events & Arguments & Entities & Relations & Average Tokens \\
\hline
Train & 8,000 & 4,368 & 14,372 & 59,915 & 24,596 & 59.5 \\
Development & 2,000 & 1,080 & 3,585 & 15,118 & 6,231 & 59.7 \\
Test & 3,000 & 1,549 & 5,130 & 22,475 & 9,425 & 59.7 \\
Total & 13,000 & 6,997 & 23,087 & 97,508 & 40,252 & 59.5 \\
\hline
\end{tabular}
\end{table}

Figure \ref{fig:dataset-distribution} shows long-tailed distributions in CMNIE. EQU and ORG dominate entity mentions; Belong, Subordinate, and Equal dominate relation annotations; and Manoeuvre is the most frequent event type. Panels (b) and (c) additionally show instances with empty relation and event annotations; these bars summarize annotation coverage and fall outside the corresponding label inventories. This pattern reflects corpus heterogeneity: relevant instances may provide background or entity context without expressing every schema layer. The instance-level distributions further show dense entity annotations and frequent relation annotations within military news instances.

\section{Benchmark Experiments}

\subsection{Experimental Setup}

The benchmark evaluates event extraction (EE), comprising trigger extraction and argument extraction \cite{peng-etal-2023-devil,zhu-etal-2024-lc4ee}, together with named entity recognition (NER) and relation extraction (RE).
Following the TextEE evaluation protocol \cite{huang-etal-2024-textee}, we report micro-averaged precision, recall, and F1 for trigger identification (TI), trigger classification (TC), argument identification (AI), and argument classification (AC), together with relaxed variants TI*, TC*, AI*, and AC*.
TI requires exact trigger span matching and ignores the Event Type.
TC additionally requires the Event Type to be correct.
AI requires exact argument span matching and ignores the Argument Role.
AC additionally requires the Argument Role to be correct.
The relaxed variants use the same label requirements but replace exact span matching with relaxed matching.
A relaxed match is counted only when the predicted span contains the complete gold span and the total extension over its left and right boundaries is no more than two tokens, with one-to-one maximum bipartite matching.
Truncated predictions and partial overlaps are excluded because they omit part of the annotated semantic unit.
We report TC and AC as the primary exact event metrics and TC* and AC* for relaxed span analysis.
For NER, a prediction is correct only if both the entity span and entity type match the gold annotation.
For RE, a prediction is correct only if the relation type and the two entity spans match the gold annotation.

The evaluated methods are grouped into supervised IE models, zero-shot LLMs, and fine-tuned LLM extraction methods.
The supervised models are DyGIE++, OneIE, and UIE \cite{lin-2020-joint,lu-2022-unified,wadden-2019-entity}.
The zero-shot LLMs include Qwen2.5 models \cite{qwen2.5}, GLM-4 \cite{glm2024chatglm}, Llama models \cite{llama3herd}, Moonshot-v1, DeepSeek-V3.1, and GPT-5.1.\footnote{Moonshot-v1: \url{https://www.moonshot.cn/}; DeepSeek-V3.1: \url{https://api-docs.deepseek.com/news/news250821}; GPT-5.1: \url{https://openai.com/index/gpt-5-1/}.}
The fine-tuned LLM extraction methods are GoLLIE, ADELIE, and KnowCoder \cite{li-etal-2024-knowcoder,qi-2024-adelie,sainz2024gollie}, with CodeLlama \cite{codellama} used as one GoLLIE backbone and Llama-3.1-8B-Instruct used for KnowCoder.
Supervised and fine-tuned models are trained on the same training split, and all methods are evaluated on the same test split; LLM-based extraction uses schema descriptions and the same label inventory as the supervised setting.

\subsection{Overall Results}

Table \ref{tab:main-results} reports the full benchmark results and reveals distinct strengths across model groups and extraction tasks.

\begin{table}[!htbp]
\caption{Full benchmark results on CMNIE. Scores are micro-averaged F1; bold indicates the best result in each metric column.}
\label{tab:main-results}
\centering
\small
\setlength{\tabcolsep}{3pt}
\begin{tabular}{l l r r r r}
\hline
Group & Model & TC & AC & NER & RE \\
\hline
\multirow{3}{*}{Supervised IE}
& DyGIE++ & 65.3 & 39.2 & 67.6 & 8.3 \\
& OneIE & 63.0 & \textbf{53.9} & 70.2 & 19.5 \\
& UIE & 47.3 & 20.4 & 64.3 & 25.1 \\
\hline
\multirow{8}{*}{Open-weight LLM}
& Qwen2.5-7B-Instruct & 37.9 & 26.3 & 39.9 & 8.0 \\
& Qwen2.5-14B-Instruct & 35.3 & 26.7 & 41.6 & 12.8 \\
& Qwen2.5-32B-Instruct & 32.7 & 29.7 & 45.0 & 16.1 \\
& Qwen2.5-72B-Instruct & 45.3 & 29.3 & 47.0 & 13.8 \\
& GLM-4-9B-Chat & 24.2 & 20.8 & 38.4 & 8.3 \\
& Llama-3.1-8B-Instruct & 23.7 & 23.1 & 26.9 & 8.3 \\
& Llama-3.1-70B-Instruct & 43.4 & 29.5 & 37.9 & 12.9 \\
& Llama-3.3-70B-Instruct & 36.1 & 29.4 & 39.9 & 15.0 \\
\hline
\multirow{3}{*}{Commercial LLM}
& Moonshot-v1-8k & 34.8 & 27.8 & 42.7 & 18.1 \\
& DeepSeek-V3.1-Chat & 49.4 & 34.4 & 46.7 & 19.2 \\
& GPT-5.1 & 39.4 & 31.8 & 46.5 & 24.2 \\
\hline
\multirow{7}{*}{Fine-tuned LLM}
& GoLLIE + CodeLlama-7B & 67.2 & 46.8 & \textbf{71.9} & 48.7 \\
& GoLLIE + Llama-3.1-8B-Instruct & 69.5 & 47.4 & 71.1 & 49.7 \\
& GoLLIE + Qwen2.5-Coder-7B & 69.1 & 45.2 & 69.9 & 49.5 \\
& GoLLIE + Qwen2.5-7B-Instruct & \textbf{70.3} & 47.9 & 71.0 & \textbf{50.2} \\
& ADELIE + Llama-3.1-8B-Instruct & 59.5 & 47.7 & 67.4 & 48.5 \\
& ADELIE + Qwen2.5-7B-Instruct & 57.7 & 45.8 & 66.7 & 46.6 \\
& KnowCoder + Llama-3.1-8B-Instruct & 46.8 & 34.6 & 68.6 & 23.6 \\
\hline
\end{tabular}
\end{table}

OneIE obtains the highest AC score (53.9), 6.0 points above the next-best result, and reaches 70.2 NER F1; GoLLIE with Qwen2.5-7B-Instruct leads TC and RE (70.3/50.2), improving RE by 25.1 points over UIE, the strongest supervised RE baseline.
This task-dependent split suggests that explicit span selection remains effective for exact arguments, while schema-conditioned generation is especially effective for relation extraction.
The three highest NER scores (71.9, 71.1, and 71.0) are closely clustered, indicating comparable NER performance among the leading systems.

KnowCoder with Llama-3.1-8B-Instruct reaches TC 46.8, AC 34.6, NER 68.6, and RE 23.6.
The strongest zero-shot results trail the overall best by 20.9 points for TC, 19.5 for AC, 24.9 for NER, and 26.0 for RE. Within this group, DeepSeek-V3.1-Chat leads TC/AC (49.4/34.4), GPT-5.1 leads RE (24.2), and Qwen2.5-72B-Instruct leads NER (47.0).
Qwen2.5 scaling is non-monotonic, indicating that model size alone provides insufficient control over schema adherence and exact boundaries.

\subsection{Exact and Relaxed Event Scores}

\begingroup
\setlength{\intextsep}{4pt}
\begin{table}[!ht]
\caption{Exact and relaxed zero-shot event extraction F1 (best in bold).}
\label{tab:span-results}
\centering
\footnotesize
\setlength{\tabcolsep}{4pt}
\begin{tabular}{l r r r r}
\hline
Model & TC & AC & TC* & AC* \\
\hline
Qwen2.5-7B-Instruct & 37.9 & 26.3 & 48.0 & 34.3 \\
Qwen2.5-14B-Instruct & 35.3 & 26.7 & 46.6 & 35.3 \\
Qwen2.5-32B-Instruct & 32.7 & 29.7 & 50.5 & 42.0 \\
Qwen2.5-72B-Instruct & 45.3 & 29.3 & 58.1 & 39.8 \\
GLM-4-9B-Chat & 24.2 & 20.8 & 29.9 & 32.0 \\
Llama-3.1-8B-Instruct & 23.7 & 23.1 & 30.0 & 28.6 \\
Llama-3.1-70B-Instruct & 43.4 & 29.5 & 64.7 & 43.7 \\
Llama-3.3-70B-Instruct & 36.1 & 29.4 & 53.2 & 35.8 \\
Moonshot-v1-8k & 34.8 & 27.8 & 46.0 & 36.2 \\
DeepSeek-V3.1-Chat & \textbf{49.4} & \textbf{34.4} & \textbf{71.3} & \textbf{43.9} \\
GPT-5.1 & 39.4 & 31.8 & 58.2 & 40.4 \\
\hline
\end{tabular}
\end{table}
\endgroup

Table \ref{tab:span-results} compares exact and relaxed event scores for zero-shot LLMs.
The gap between exact and relaxed F1 is large for nearly every model.
DeepSeek-V3.1-Chat reaches 71.3 TC* but 49.4 TC.
Llama-3.1-70B-Instruct reaches 64.7 TC* but 43.4 TC.
The same pattern appears in AC, where relaxed scores are consistently higher than exact scores.

Relaxation increases TC by 21.9 points for DeepSeek-V3.1-Chat and 21.3 for Llama-3.1-70B-Instruct; the largest AC increase is 12.3 points for Qwen2.5-32B-Instruct.
DeepSeek-V3.1-Chat reaches 71.3 TC*, close to the best exact TC of a fine-tuned system, highlighting boundary precision as a major source of zero-shot error.
Exact spans determine which phrase enters a structured record: overextended spans may add irrelevant modifiers, while shortened spans may omit equipment models, organization names, or location components.

\section{Discussion}

\subsection{Challenges and Error Analysis}

CMNIE combines skewed event distributions, dense entities, and frequent relations. Models must distinguish background instances with empty event annotations, learn sparse types, and recover precise boundaries; many relations also appear in noun phrases or appositions without an explicit predicate.

As a qualitative case study, we manually inspected outputs from Qwen2.5-7B-Instruct across event extraction, NER, and RE.
Event errors include missed triggers, false predictions over background descriptions, and role confusions; NER errors concentrate on boundaries of equipment models, organizations, and locations; and RE errors arise from inaccurate entity spans and implicit relation expressions.
Because RE requires two exact entity endpoints, a single NER boundary error can invalidate an otherwise plausible relation prediction, compounding entity and relation errors.
These observations align with the relaxed--exact gap and low zero-shot RE scores. They motivate boundary-aware decoding, schema-constrained generation, and tighter entity--relation coupling; OneIE's AC lead also shows the value of explicit span selection.

\subsection{Data Availability, Limitations, and Ethical Considerations}

CMNIE covers public Chinese military news from the selected sources and period. Its schema captures a defined set of concepts; some event types remain sparse, and the current release covers Chinese only. Dictionary retrieval may favor explicit, frequent event expressions and provides limited coverage of implicit or atypical events outside the dictionary. The dataset and code are available at \url{https://github.com/Sulfur-ylide/CMNIE}.

The dataset was collected solely from publicly accessible news sources. Military IE has dual-use potential, including automated monitoring and operational analysis. Users should follow source copyright and redistribution conditions, applicable legal and ethical requirements, and avoid harmful deployment; the release documents provenance and distributes permitted materials.

\section{Conclusion}

This paper presents CMNIE, a Chinese military news benchmark that aligns events, arguments, entities, and relations under a unified schema. Experiments reveal complementary strengths: explicit span modeling remains effective for exact arguments, while fine-tuned schema-conditioned generation achieves the strongest relation extraction results. Exact--relaxed gaps further establish boundary precision as a central zero-shot challenge. CMNIE supports research on schema adherence, span-sensitive extraction, and joint structured prediction in specialized Chinese news.


\bibliographystyle{splncs04}
\bibliography{cas-refs}

\end{document}